\documentclass{article}

\usepackage{PRIMEarxiv}

\usepackage[utf8]{inputenc} 
\usepackage[T1]{fontenc}    
\usepackage{hyperref}       
\usepackage{url}            
\usepackage{booktabs}       
\usepackage{amsfonts}       
\usepackage{nicefrac}       
\usepackage{microtype}      
\usepackage{lipsum}
\usepackage{fancyhdr}       
\usepackage{graphicx}       
\usepackage{subcaption} 
\usepackage{appendix} 
\graphicspath{{media/}}     

\title{Rethinking the adaptive relationship between Encoder Layers and Decoder Layers}

\author{Yubo Song \\ \texttt{yubsog@163.com}}

\begin{document}
\maketitle

\begin{abstract}
This article explores the adaptive relationship between Encoder Layers and Decoder Layers using the SOTA model Helsinki-NLP/opus-mt-de-en, which translates German to English. The specific method involves introducing a bias-free fully connected layer between the Encoder and Decoder, with different initializations of the layer's weights, and observing the outcomes of fine-tuning versus retraining. Four experiments were conducted in total. The results suggest that directly modifying the pre-trained model structure for fine-tuning yields suboptimal performance. However, upon observing the outcomes of the experiments with retraining, this structural adjustment shows significant potential.
\end{abstract}

\keywords{Transformer \and Decoder Layers \and Encoder Layers}

\section{Introduction}
In the field of machine learning, using pre-trained models to perform specific tasks is a common practice. Typically, this involves fine-tuning the pre-trained model on a specific dataset through iterative adjustments without modifying the model structure. This article focuses on the state-of-the-art (SOTA) machine translation model Helsinki-NLP/opus-mt-de-en, which translates German to English, to explore the adaptive relationship between Encoder Layers and Decoder Layers by introducing a bias-free fully connected layer. Additionally, the study investigates the effects of modifying the pre-trained model structure during fine-tuning.

Four experiments were conducted by introducing a bias-free fully connected layer between the Encoder and Decoder Layers:
\begin{itemize}
    \item Using original pre-trained model weights and initializing the fully connected layer weights to maintain the original connections, where each Decoder Layer's input is from the 6th Encoder Layer. Through fine-tuning, these weights adapt towards optimal configurations.
    \item Fine-tuning the original pre-trained model as a baseline experiment.
    \item Initializing the fully connected layer weights with Granularity Consistent Attention.
    \item Initializing all layer weights with a variance of 0 and mean of 1, with the fully connected layer between Encoder and Decoder initialized to maintain original connections.
\end{itemize}
Analysis of the experimental results yielded important insights: Directly modifying the pre-trained model structure during fine-tuning might lead to performance degradation, potentially due to the pre-trained model's weights being optimized for its original structure. However, results from experiments involving retraining suggest that appropriate modifications to the model structure have significant potential to enhance the performance and stability of machine translation.

This research aims to contemplate the adaptive relationship between Encoder Layers and Decoder Layers, alongside experimenting with modifications to the pre-trained model structure, offering valuable insights and directions for further development and optimization in the field of natural language processing. Reproducible Code \href{https://github.com/parallelsucc/opus}{GitHub}.

The rest of the paper is organized as follows: Section \ref{sec:related work} introduces related work, Section \ref{sec:approach} describes the experimental methodology of this study, Section \ref{sec:Experiment and Analysis} presents the experiments and analysis, and Section \ref{summary} provides a summary of the entire paper.

\section{Related work}
\label{sec:related work}

In 2017, Vaswani et al \cite{vaswani2017attention}. introduced the transformer architecture, which led to the development of transformer-based Encoder-Decoder models.

Liu et al \cite{Liu2020RethinkingAI}.  conducted a series of comparative experiments on (a) Granularity Consistent Attention; (b) Granularity Parallel Attention; (c) Fine-Grained Attention; (d) Full Matching Attention; (e) Adaptive Matching Attention. The experimental results demonstrated that Granularity Consistent Attention (GCA) outperformed other approaches on Transformer and DynamicConv architectures.

Tian et al \cite{TIAN}. integrated the encoder as part of the decoder, enabling certain modules of the decoder to share parameters with the encoder. They used a gating mechanism to filter key information from the input sequence, which improved the training and inference speed of text summarization tasks, while also enhancing the accuracy and fluency of generated summaries.

Devlin et al \cite{devlin-etal-2019-bert}. leveraged the self-attention mechanism of the Transformer while considering both left and right context information for each word. This bidirectional context modeling allows for a more comprehensive representation of each word's semantic features within the entire sentence. This bidirectional context modeling contributes to enhancing language understanding and performance in downstream tasks.

\section{Approach}
\label{sec:approach}

\begin{figure}[h] 
    \centering
    \begin{subfigure}[h]{0.44\textwidth} 
        \centering
        \includegraphics[width=\textwidth]{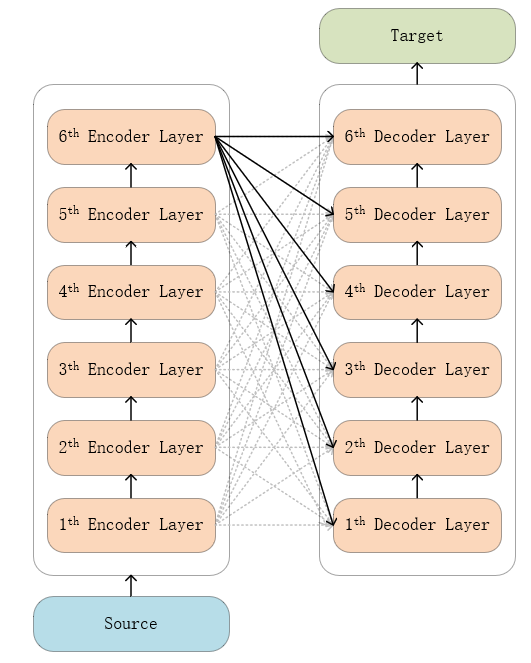} 
        \caption{Initialized as original structure} 
        \label{fig:Initialized as original structure} 
    \end{subfigure}
    \hspace{2em} 
    \begin{subfigure}[h]{0.49\textwidth} 
        \centering
        \includegraphics[width=\textwidth]{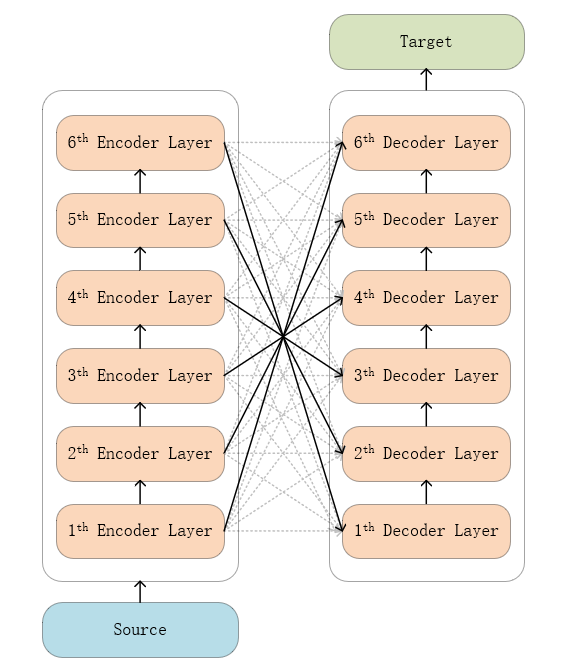} 
        \caption{Initialized as GCA} 
        \label{fig:Initialized as GCA} 
    \end{subfigure}
    \caption{Weight Initialization Methods} 
    \label{fig:Weight Initialization Methods} 
\end{figure}

The paper explores the relationship between Encoder Layers and Decoder Layers by introducing a fully connected layer between them and experimenting with different weight initialization methods. As depicted in Figure \ref{fig:Weight Initialization Methods}, this approach effectively adjusts the interaction of information between the encoder and decoder, thereby impacting the model's performance and learning capabilities.

Specifically, the introduction of the fully connected layer allows for more flexible information propagation and transformation. By inserting a fully connected layer between the output of the Encoder and the input of the Decoder, the hidden states generated by the Encoder undergo a linear transformation to adapt to the input requirements and task demands of the Decoder. This linear transformation, governed by different weight initialization methods, explores various relationships between Encoder and Decoder layers, influencing the model's representation learning and feature extraction capabilities.

Figure \ref{fig:Weight Initialization Methods} illustrates the schematic diagram of this fully connected layer structure, where the output of the Encoder is passed through the fully connected layer before being used as input for the Decoder. This design, while preserving the original information, enables exploration and optimization of the flow of information between the Encoder and Decoder through adjustments in weight initialization, thereby enhancing the model's performance and generalization capabilities.

Figure \ref{fig:Initialized as original structure} represents initialization as original structure, and Figure \ref{fig:Initialized as GCA} represents initialization as GCA. For different Encoder Layers, higher layers correspond to more abstract representations. Higher layers focus more on global information, while lower layers emphasize local information.

In Figure \ref{fig:Initialized as original structure}, each input to every Decoder Layer comes directly from the final layer output of the Encoder. This simplified connection process allows the Decoder to directly access all hidden states from the final layer of the Encoder, providing rich source language representations. However, this approach has certain drawbacks. Since each Decoder layer can only perceive global information from the entire input sequence, the model may tend to overlook local details or subtle semantic features within the input sequence. This global connection method can lead to suboptimal performance when dealing with complex language structures or long sentences because the Decoder lacks detailed knowledge of local information.

In Figure \ref{fig:Initialized as GCA}, the connection between Encoder and Decoder employs a Global Context Attention (GCA) mechanism. This connection strategy enables each decoding step to be viewed as a progressively refining process, where abstract high-level semantic information gradually transforms into specific low-level word sequences. GCA preserves the overall integrity of source language information while allowing the Decoder to progressively process and understand the input information. Specifically, the GCA mechanism assists Decoder layers in sequentially processing input information to better capture local features and semantic relationships within the source language. Compared to the straightforward global connection method, GCA enhances the model's ability to handle complex input sequences, improving performance and generalization when dealing with long sentences or text of high semantic complexity.

\section{Experiment and Analysis}
\label{sec:Experiment and Analysis}

This paper takes the state-of-the-art (SOTA) machine translation model Helsinki-NLP/opus-mt-de-en, which translates German to English, as an example. It introduces a fully connected layer between the Encoder and Decoder to observe the adaptive relationship between Encoder Layers and Decoder Layers using different weight initialization methods. Specifically, the paper preserves hidden states, where the output of each Encoder Layer serves as input to the fully connected layer. The output of the fully connected layer then becomes the input to each Decoder Layer, maintaining the original input shape of the Decoder Layer. For Helsinki-NLP/opus-mt-de-en, the shape of the fully connected layer is 3072x512, with a total of 6 fully connected layers.

\subsection{Details}
\begin{figure}[htbp]
    \centering
    \includegraphics[width=1\textwidth]{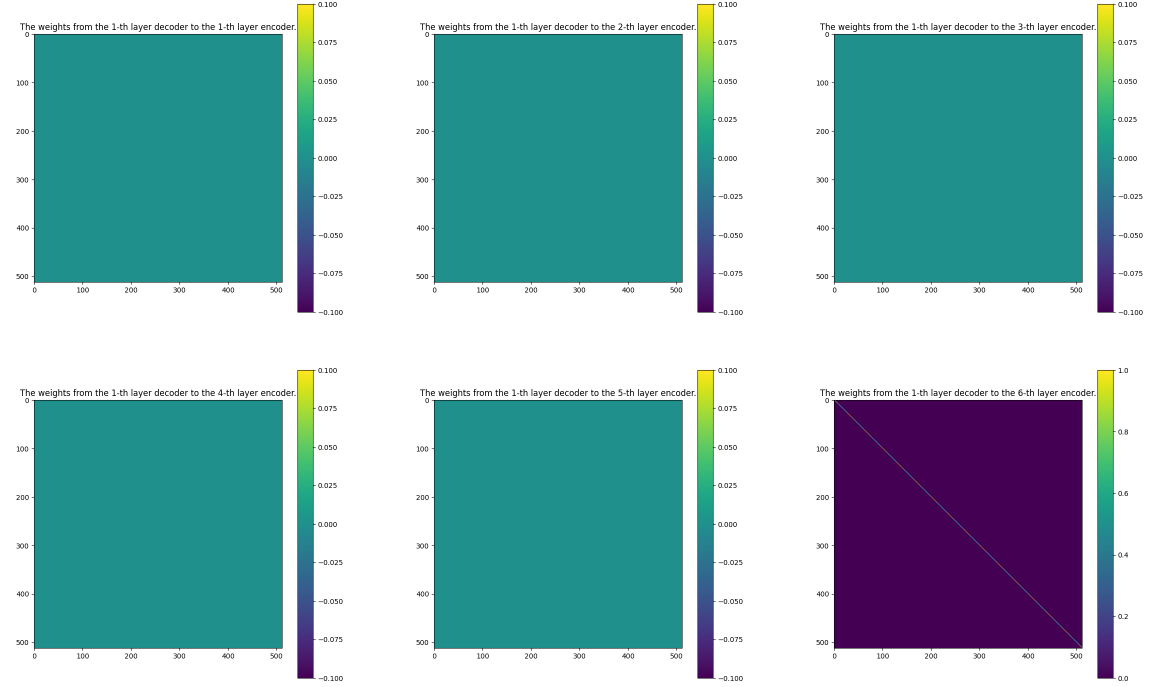}
    \caption{Initialized to the weight of the original connection method}
    \label{fig:Initialized to the weight of the original connection method}
\end{figure}

\begin{figure}[htbp]
    \centering
    \includegraphics[width=0.96\textwidth]{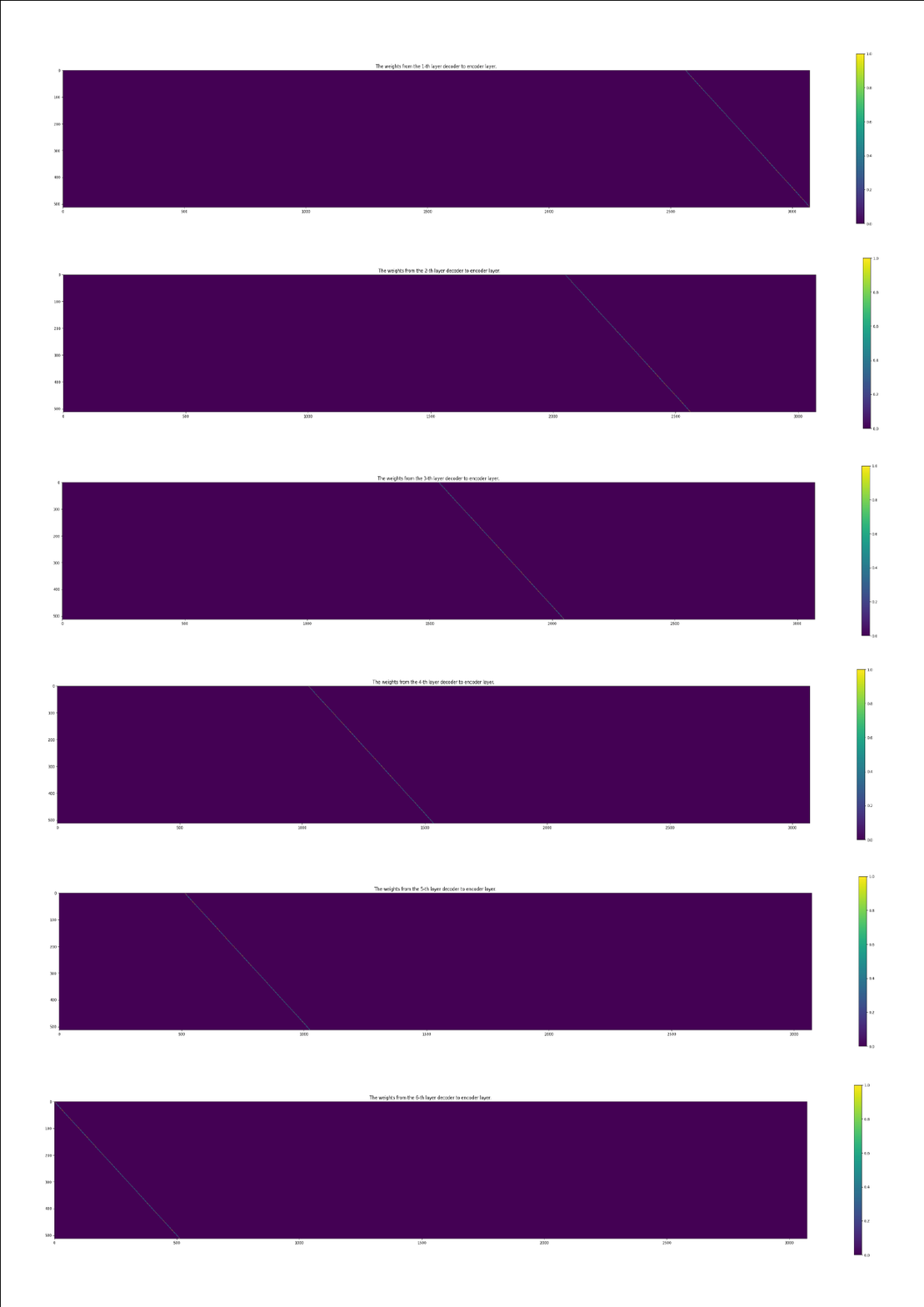}
    \caption{Initialized to the weight of GCA method}
    \label{fig:Initialized to the weight of GCA method}
\end{figure}

The dataset used in this study is the wmt16/de-en dataset. The training set consists of the first 1 million entries from the training data in wmt16, the validation set is from the wmt16 validation dataset, and the test set is from the wmt16 test dataset. The fine-tuning is conducted for 1 epoch, while the retraining is carried out for 1 epoch as well. The weight initialization of the fully connected layer is depicted in Figure \ref{fig:Initialized to the weight of the original connection method} \ref{fig:Initialized to the weight of GCA method}.

Figure \ref{fig:Initialized to the weight of the original connection method} represents initialization as the original connection method, where each fully connected layer is initialized with identical weights. Specifically, each Decoder Layer has unit matrix weights for the last Encoder Layer and zero weights for other Encoder Layers.

Figure \ref{fig:Initialized to the weight of GCA method} represents initialization as the GCA connection method. In this approach, the weights are initialized such that the first Decoder Layer has unit matrix weights for the last Encoder Layer, the second Decoder Layer has unit matrix weights for the second-to-last Encoder Layer, and so forth, following this pattern.


\subsection{Results and Analysis}
\begin{figure}[htbp] 
    \centering

    \begin{subfigure}[htbp]{0.45\textwidth}
        \centering
        \includegraphics[width=\textwidth]{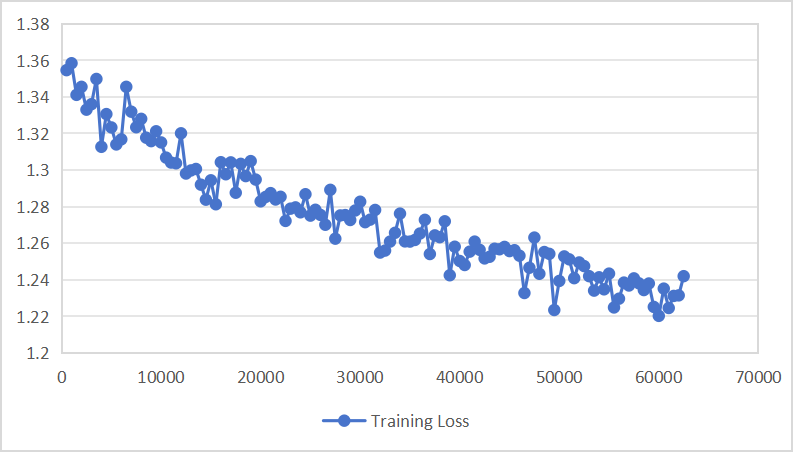} 
        \caption{Initialized as original connection method - Fine-tuning loss}
        \label{fig:Initialized as original connection method - Fine-tuning loss}
    \end{subfigure}
    \hfill
    \begin{subfigure}[htbp]{0.49\textwidth}
        \centering
        \includegraphics[width=\textwidth]{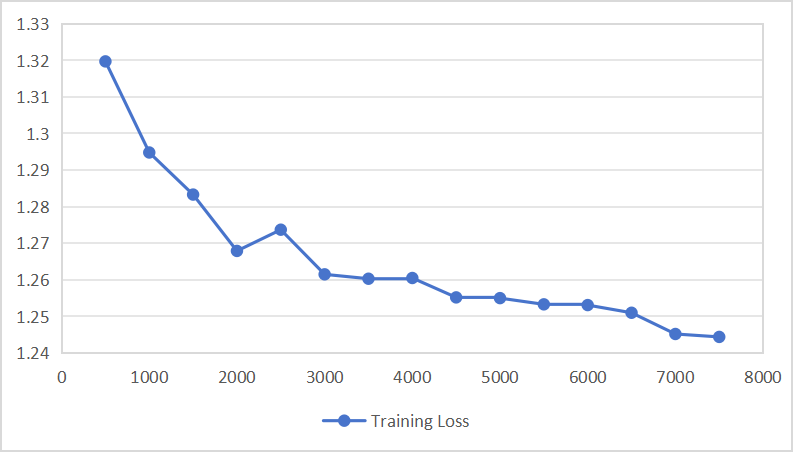} 
        \caption{Direct fine-tuning loss}
        \label{fig:Direct fine-tuning loss}
    \end{subfigure}

    \vspace{1cm} 

    \begin{subfigure}[htbp]{0.49\textwidth}
        \centering
        \includegraphics[width=\textwidth]{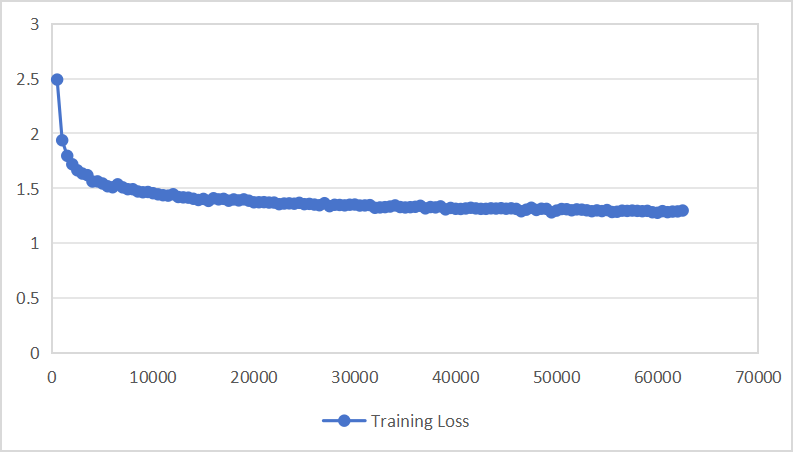} 
        \caption{Initialized as GCA - Fine-tuning loss}
        \label{fig:Initialized as GCA - Fine-tuning loss}
    \end{subfigure}
    \hfill
    \begin{subfigure}[htbp]{0.45\textwidth}
        \centering
        \includegraphics[width=\textwidth]{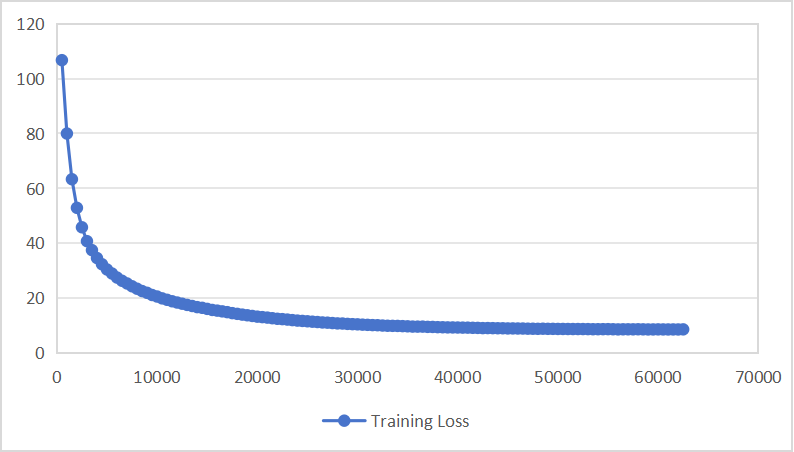} 
        \caption{Initialized as original connection method - Retraining loss}
        \label{fig:Initialized as original connection method - Retraining loss}
    \end{subfigure}

    \caption{Training Loss of different experiments}
    \label{fig:Training Loss of different experiments}
\end{figure}

The experimental training loss for different approaches is illustrated in Figure \ref{fig:Training Loss of different experiments}.

As shown in Figure \ref{fig:Training Loss of different experiments}, when initialized with the original connection method and subjected to fine-tuning, the training loss exhibits significant fluctuations and fails to converge within 1 epoch. The authors suggest that this behavior may reflect the model's adaptation to the fully connected layer. Increasing the amount of training data and epochs could potentially yield better performance. On the other hand, direct fine-tuning shows smaller fluctuations, which is noteworthy considering the larger batch size of 128 compared to 16 used in other experiments. This unusual fluctuation warrants further investigation.

In the case of initializing with GCA and conducting fine-tuning, the higher training loss is attributed to the model's structural complexity, indicating a need for more training data to improve performance.

The initialization with the original connection method aims to explore better information transmission between Encoder and Decoder. Despite the higher training loss due to limited training data, the distribution of weights in the fully connected layer remains informative. As shown in Appendix \ref{A Appendix}, the retraining with the original connection method did not exhibit significant differences in the fully connected layer weights due to the limited amount of training data. However, after training, the weights originally focusing on the Last Encoder Layer became more distributed and similar to other Encoder Layers, demonstrating the positive influence of Encoder Layers on Decoder Layers.

As shown in Appendix \ref{B Appendix} and Figure \ref{fig:Initialized to the weight of GCA method}, initializing the fully connected layer weights with GCA and subjecting them to fine-tuning resulted in substantial changes. The weights transitioned from the original GCA connection method to a fully connected structure, indicating that each Decoder Layer attempts to access information from all Encoder Layers in varying proportions.

\begin{table}[h]
\caption{\textbf{Results of 4 Experiments}}
\centering
\begin{tabular}{lcc}
\toprule
 &Evaluate Loss&BLEU \\
\midrule
Initialized with original connection, fine-tuned&1.336&33.5984 \\
Direct fine-tuning&\textbf{1.1815}&\textbf{36.4331} \\
Initialized with GCA, fine-tuned&1.4602&32.1451 \\
Initialized with GCA, retrained&11.0151&0.0003 \\
\bottomrule
\end{tabular}
\end{table}

Due to limited resources, the study did not utilize the entire training dataset, and each experiment was limited to 1 epoch. This limitation significantly impacted the retraining experiment, but it serves as a prospective insight. By observing the training loss, this structure shows significant potential.

The experiment of initializing with the original connection method and fine-tuning was expected to perform at least as well as direct fine-tuning, but the results did not meet this expectation. The authors speculate that this discrepancy may be attributed to overfitting.

Due to the pretrained model weights being trained on the original structure, they are not well-suited for the modified GCA structure. Therefore, initializing the fully connected layer weights as GCA and then fine-tuning did not yield as good results as direct fine-tuning. However, increasing the amount of training data and training epochs may lead to improved outcomes.

The differences in Evaluate Loss and BLEU between initialization with the original connection method and GCA are not significant, but the original connection method is more compatible with the pretrained model. Therefore, increasing the amount of training data or epochs may enhance the performance of models initialized with the GCA approach. This underscores that adapting the fully connected layer weights using GCA for adaptive adjustments can enable the model to better capture local features and semantic relationships within the source language.

\section{Summary}
\label{summary}
This paper explores better information transmission between Encoder and Decoder by observing weight changes with the inclusion of a fully connected layer. The weights are initialized using both the original connection method and the GCA method. After fine-tuning with 1 million data samples and 1 epoch, there was little difference between the two approaches. However, since the pretrained model weights are based on the original connection method, the GCA connection method does not fully adapt to the pretrained model weights. Nonetheless, under limited training data and epochs for fine-tuning, the GCA connection method still performs well.

Moreover, based on the experimental results from fine-tuning with GCA initialization and retraining with original connection method initialization, it is concluded that each Encoder Layer positively contributes to Decoder Layer operations.

\bibliographystyle{unsrt}  
\bibliography{references}  

\appendix 

\section{Appendix}
\label{A Appendix}

\includegraphics[width=1\textwidth]{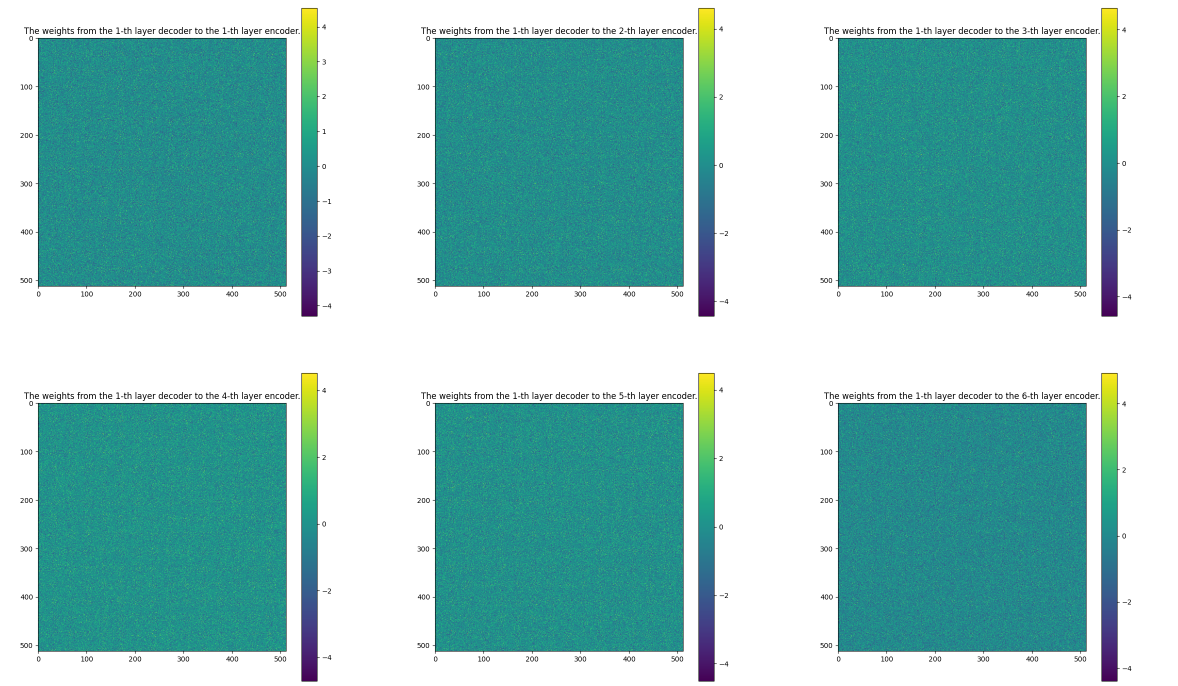}
\includegraphics[width=1\textwidth]{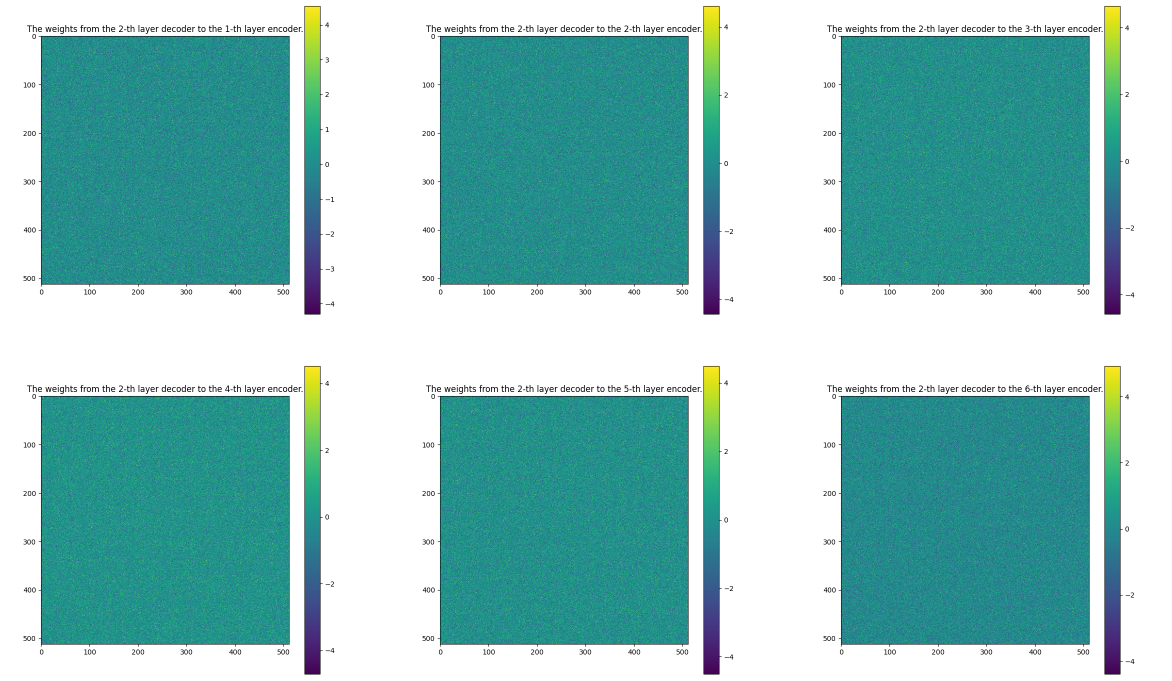}
\includegraphics[width=1\textwidth]{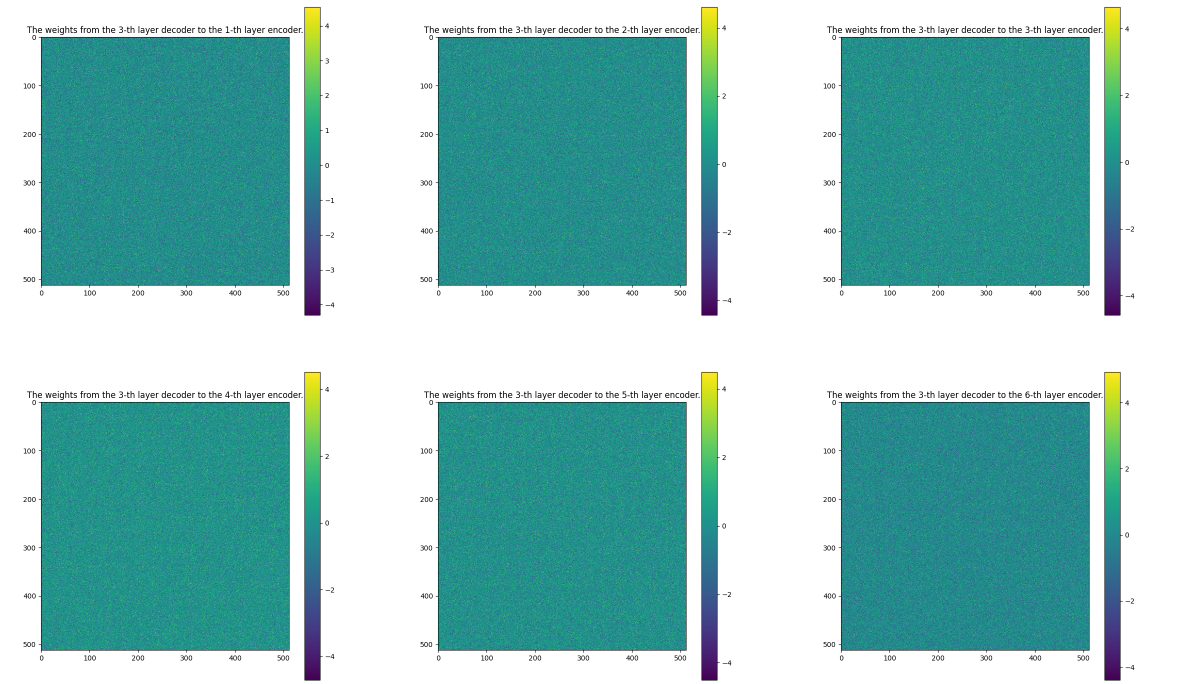}
\includegraphics[width=1\textwidth]{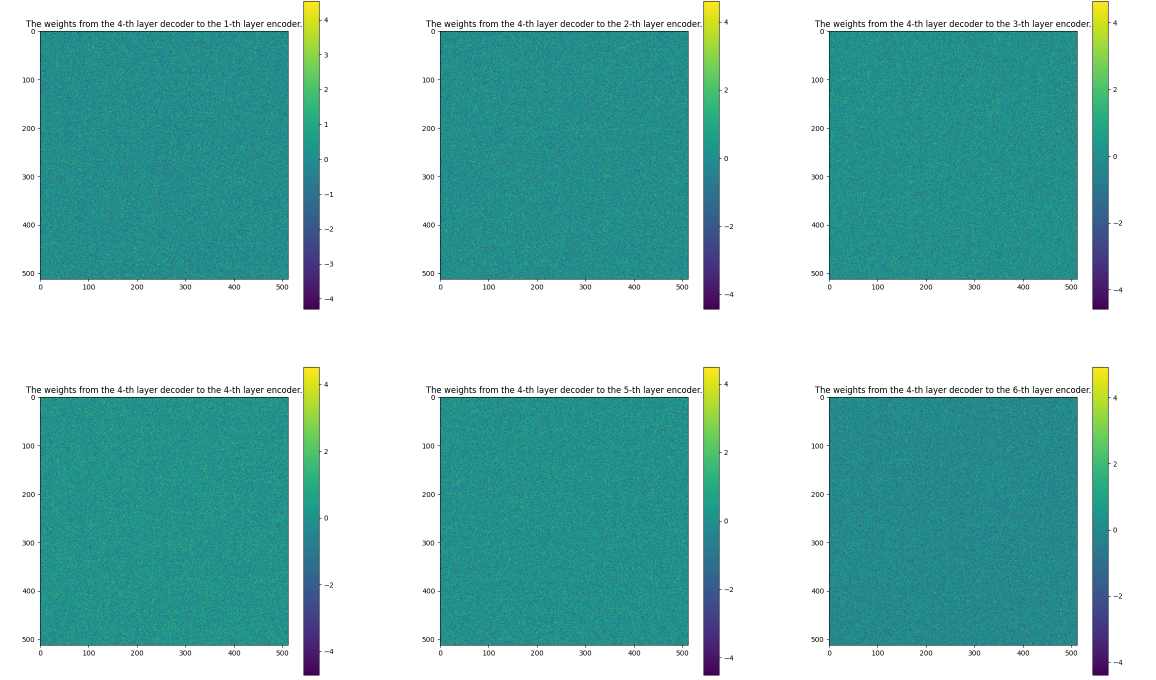}
\includegraphics[width=1\textwidth]{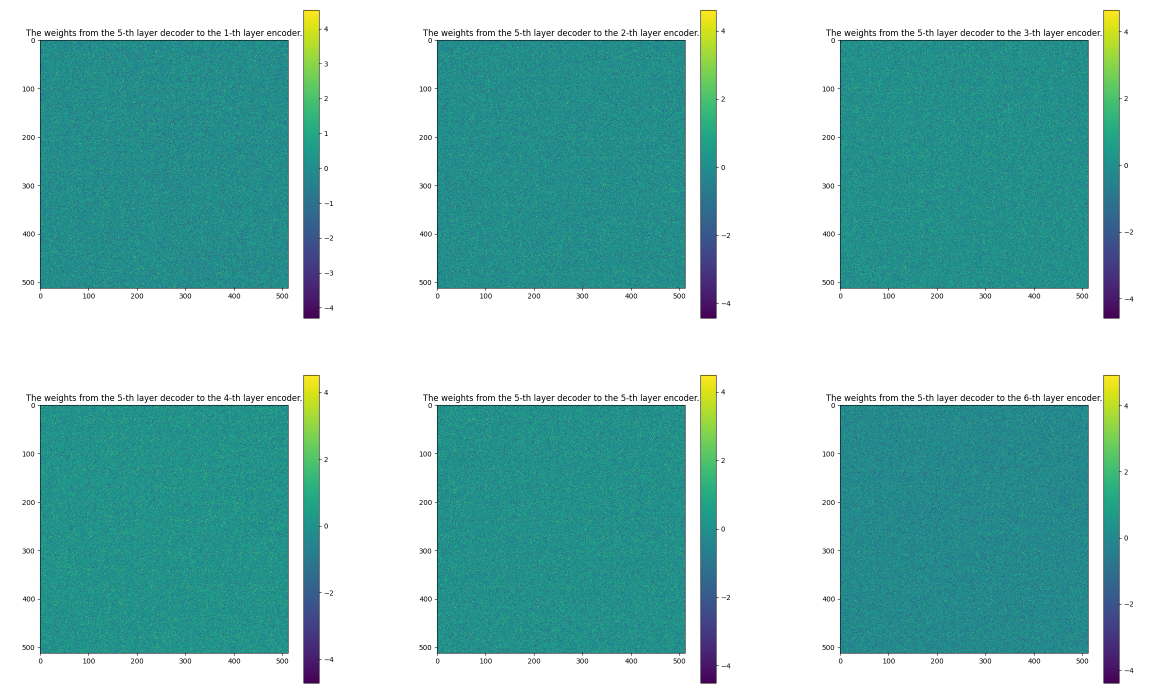}
\includegraphics[width=1\textwidth]{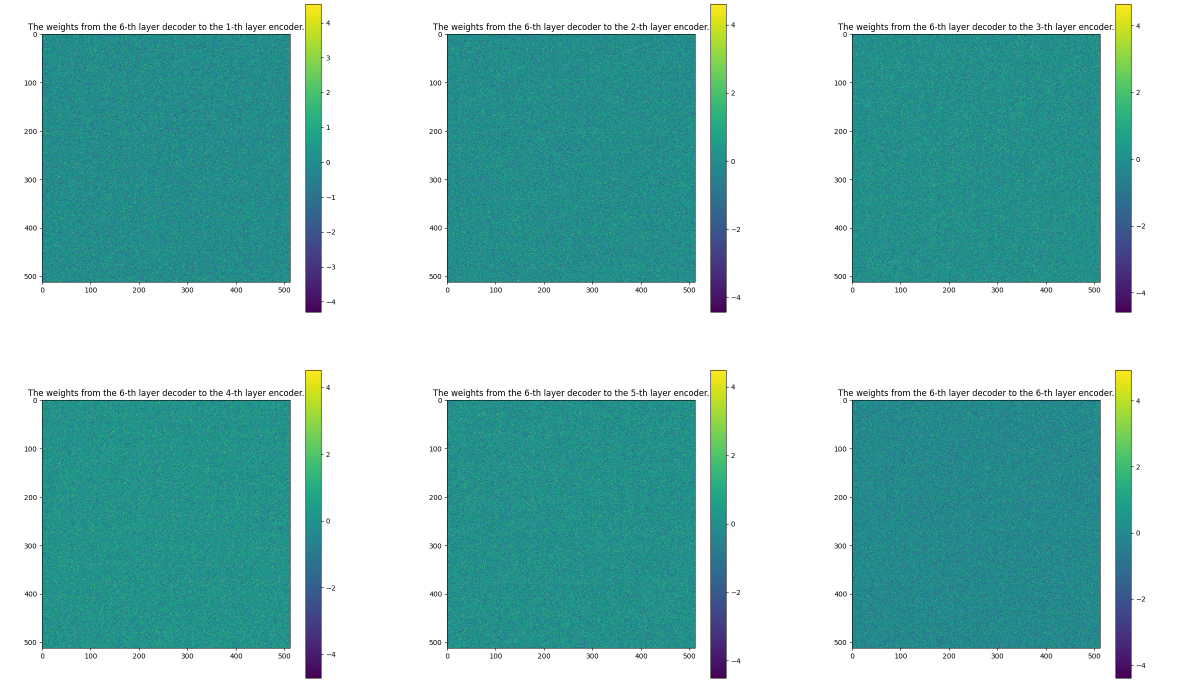}

\section{Appendix}
\label{B Appendix}

\includegraphics[width=1\textwidth]{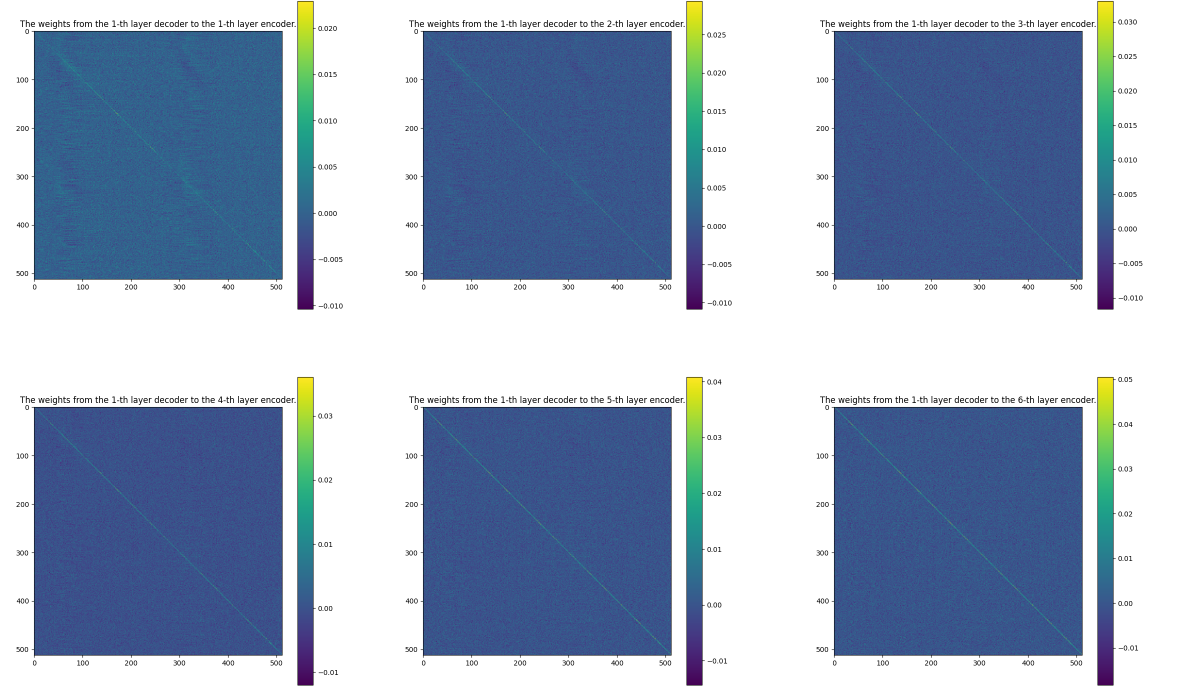}
\includegraphics[width=1\textwidth]{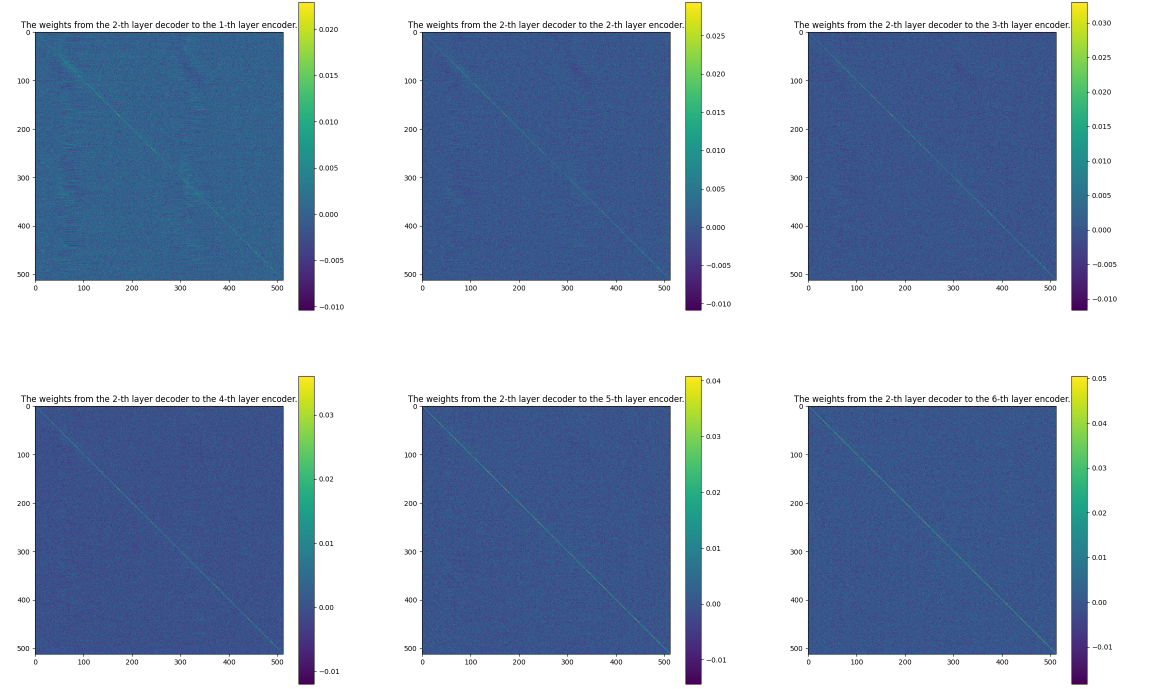}
\includegraphics[width=1\textwidth]{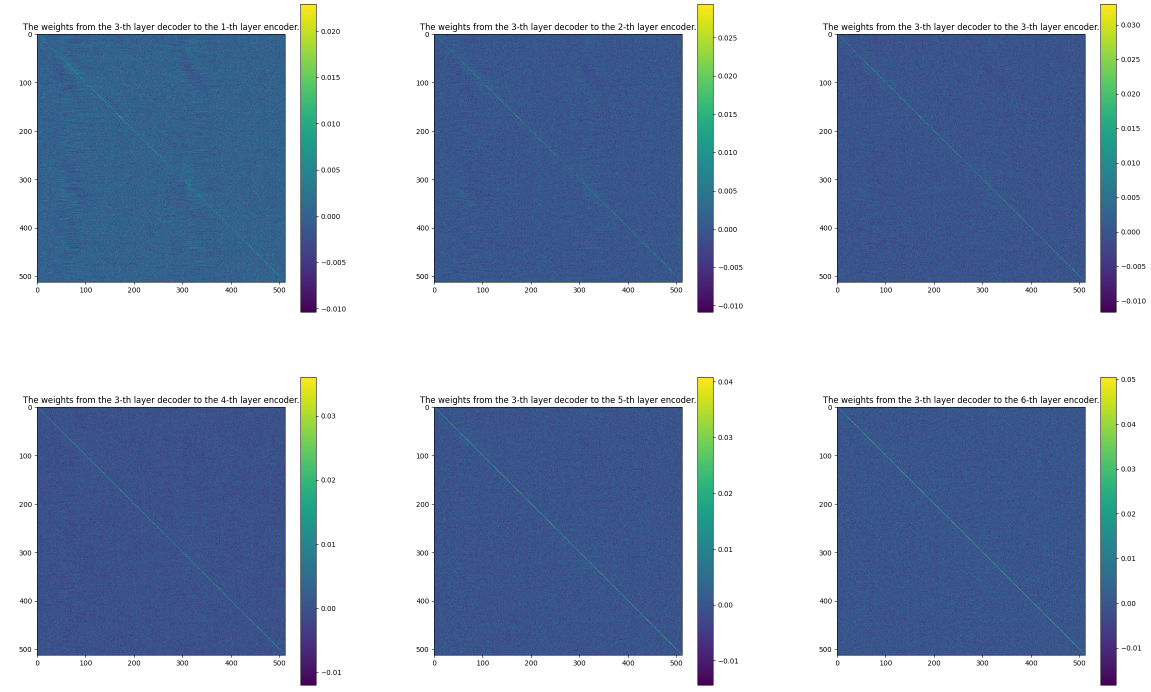}
\includegraphics[width=1\textwidth]{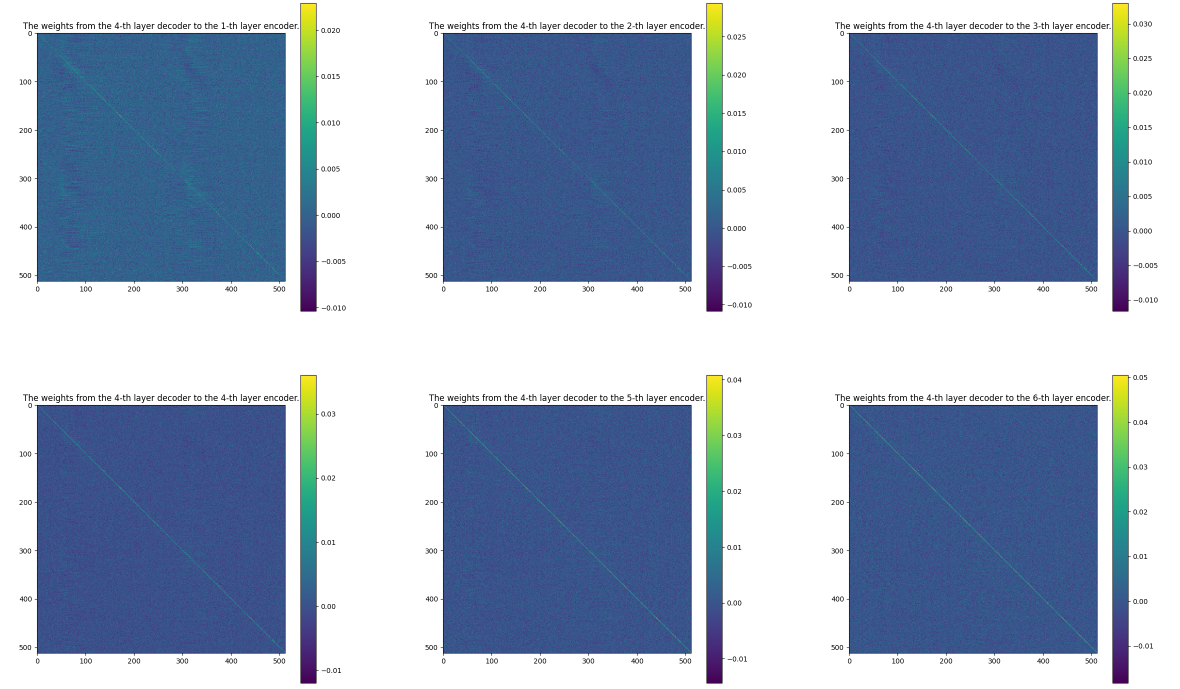}
\includegraphics[width=1\textwidth]{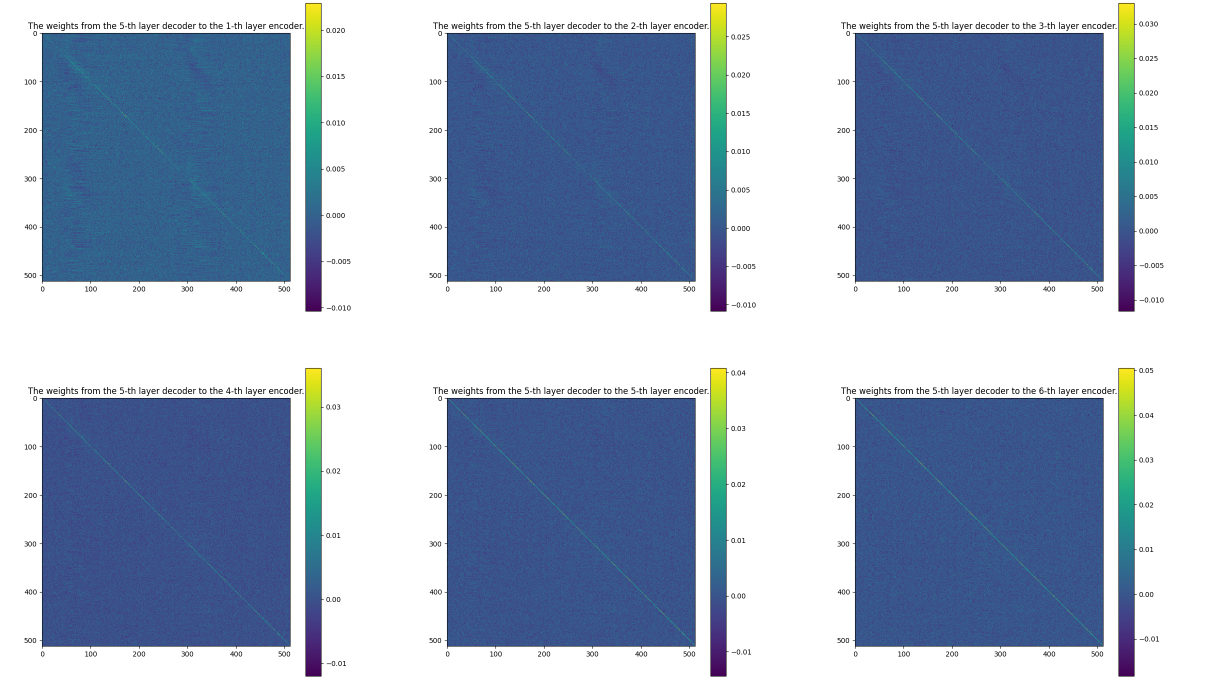}
\includegraphics[width=1\textwidth]{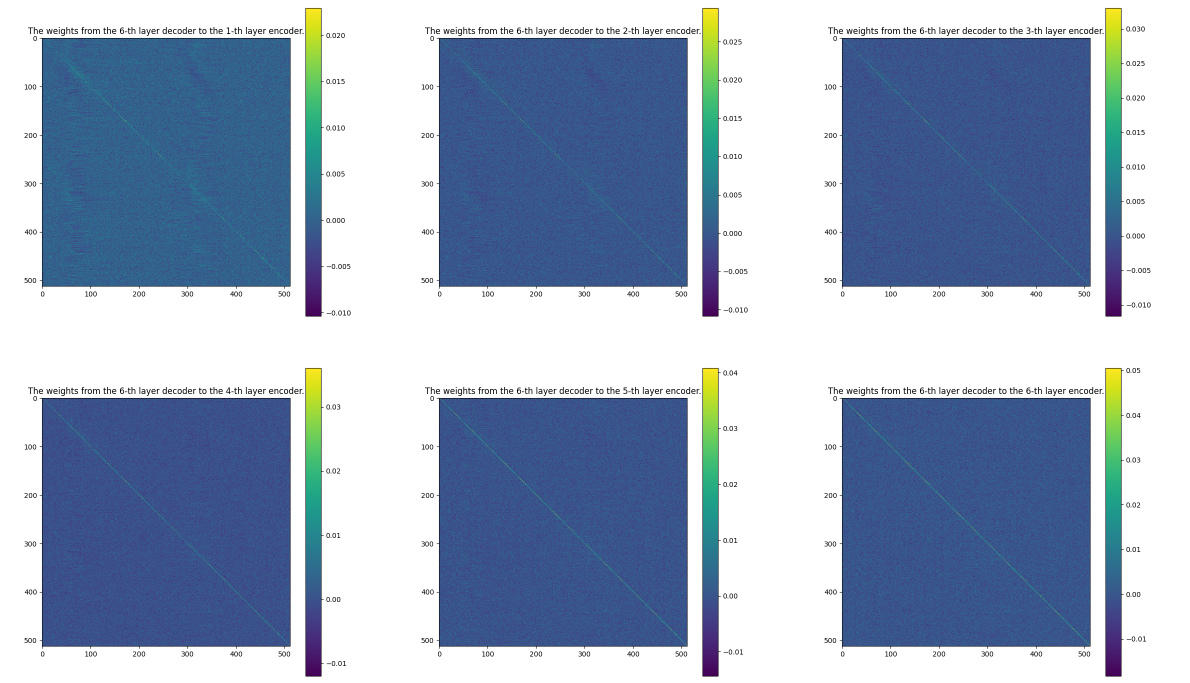}

\end{document}